\documentclass{article}

\usepackage[english]{babel}
\usepackage{authblk}
\usepackage[letterpaper,top=2cm,bottom=2cm,left=3cm,right=3cm,marginparwidth=1.75cm]{geometry}

\usepackage{amsmath}
\usepackage{graphicx}
\usepackage{hyperref}
\hypersetup{colorlinks, allcolors=black}

\title{Federated Data Model}
\author[1]{Xiao Chen}
\author[2]{Shunan Zhang}
\author[1]{Eric Zhang Chen}
\author[1]{Yikang Liu}
\author[1]{Lin Zhao}
\author[1]{Terrence Chen}
\author[1]{Shanhui Sun}
\affil[1]{United Imaging Intelligence, Burlington, MA, USA}
\affil[2]{United Imaging Intelligence, Shanghai, China}
\date{}

\begin{document}
\maketitle

\begin{abstract}
In artificial intelligence (AI), especially deep learning, data diversity and volume play a pivotal role in model development. However, training a robust deep learning model often faces challenges due to data privacy, regulations, and the difficulty of sharing data between different locations, especially for medical applications. To address this, we developed a method called the Federated Data Model (FDM). This method uses diffusion models to learn the characteristics of data at one site and then creates synthetic data that can be used at another site without sharing the actual data. We tested this approach with a medical image segmentation task, focusing on cardiac magnetic resonance images from different hospitals. Our results show that models trained with this method perform well both on the data they were originally trained on and on data from other sites. This approach offers a promising way to train accurate and privacy-respecting AI models across different locations.
\end{abstract}

% \section{outline}
% aim: solve the distributed data issue with diffusion model to achieve "federated" data 
% introduction and background:
% 1. distributed data issue: data privacy, regulation, communication; different data distribution, domain difference.  
% 2. current methods: federated learning, adapation: unsupervised or weakly supervised domain transfer, meta learning. finetune
% 3. generative ai: diffusion model for data synthesis, and privacy
% methods:
% 1. problem statement: segmentation
% 2. experiment setup
% 3. diffusion model
% 4. segmentation model
% results:
% 1. local train and local test ; remote train and local test; remote train and local finetune; synthesis and local train and local test (both sites); synthesis train and remote test
% discussion and conclusion:

\section{Introduction}

The advent of artificial intelligence (AI), especially deep learning (DL), has been significantly propelled by the voluminous data available for training neural networks. The success of these networks hinges on the quantity, quality, and diversity of the training data. Although primary models can be developed using locally available limited datasets, achieving robustness in real-world applications necessitates models capable of handling varied situations. This is particularly challenging when models trained in lab settings are deployed in real-world scenarios, often underperforming due to their inability to generalize across different data distributions. This issue, known as "domain shift," is particularly problematic in fields like medicine and industry, where data variability across different sources is compounded by differing collection protocols, vendor equipment, and underlying characteristics of the data.

Traditionally, one approach to mitigate domain shift involves training neural networks on comprehensive datasets, encompassing a wide variety of data instances. This method has shown promise, as demonstrated by recent breakthroughs in natural language processing \cite{chatgpt} and image processing technologies \cite{dalle}. However, the feasibility of collecting such extensive datasets is constrained in domains like healthcare, where data privacy concerns and regulatory restrictions limit data sharing and aggregation.

Federated learning (FL) \cite{fl} has emerged as a strategy to overcome these barriers, allowing models to learn from decentralized data without actual data transfer. Despite its promise, FL faces challenges related to communication overheads and scalability. Alternatively, model adaptation techniques, such as meta-learning \cite{meta} and fine-tuning, adjust pre-trained models to new data distributions but often require direct access to diverse datasets, which may not always be feasible.

In light of these challenges, our study introduces a versatile approach centered on data adaptation and generation through the use of diffusion models. This methodology transcends the conventional reliance on direct data sharing or model adaptation. By leveraging diffusion models \cite{ddpm} to learn data distributions at any given local site and then generating synthetic data representative of these distributions, we enable a scalable solution to the challenges of data privacy, regulation, and transfer. This synthetic data can then be shared with any number of other sites, thereby facilitating the training of robust models capable of handling domain shifts without the direct exchange of sensitive or regulated data. We termed the approach as "Federated Data Model (FDM)", as the data model is shared rather than the data itself. We validated our approach through a realistic medical imaging segmentation task involving two hospitals. The results demonstrate the feasibility and efficacy of applying our method across multiple sites, offering a promising avenue for developing robust models in privacy-sensitive and regulated environments.

\section{Methods}
% In this section we first introduce the strategy proposed here. We then describe the experiment setting which includes the medical image segmentation task we used as an example. this is followed by a bit more detail on the diffusion model and segmentation model. It is worth pointing out that we use a two-site scenario as an example, but our method can be readily extended to multiple sites. It is also worth pointing out that our method is not limited to a specific downstream task or diffusion models.
This section outlines our proposed strategy, details the experimental setup—including the medical image segmentation task used as an exemplar—and elaborates on the diffusion and segmentation models employed. While our experiments are demonstrated in a two-site scenario for simplicity, it is crucial to note that our methodology is designed to be scalable to multiple sites. Additionally, our approach is versatile, applicable not only to different downstream tasks but also adaptable with various diffusion model architectures.
\begin{figure}
\centering
\includegraphics[width=0.75\linewidth]{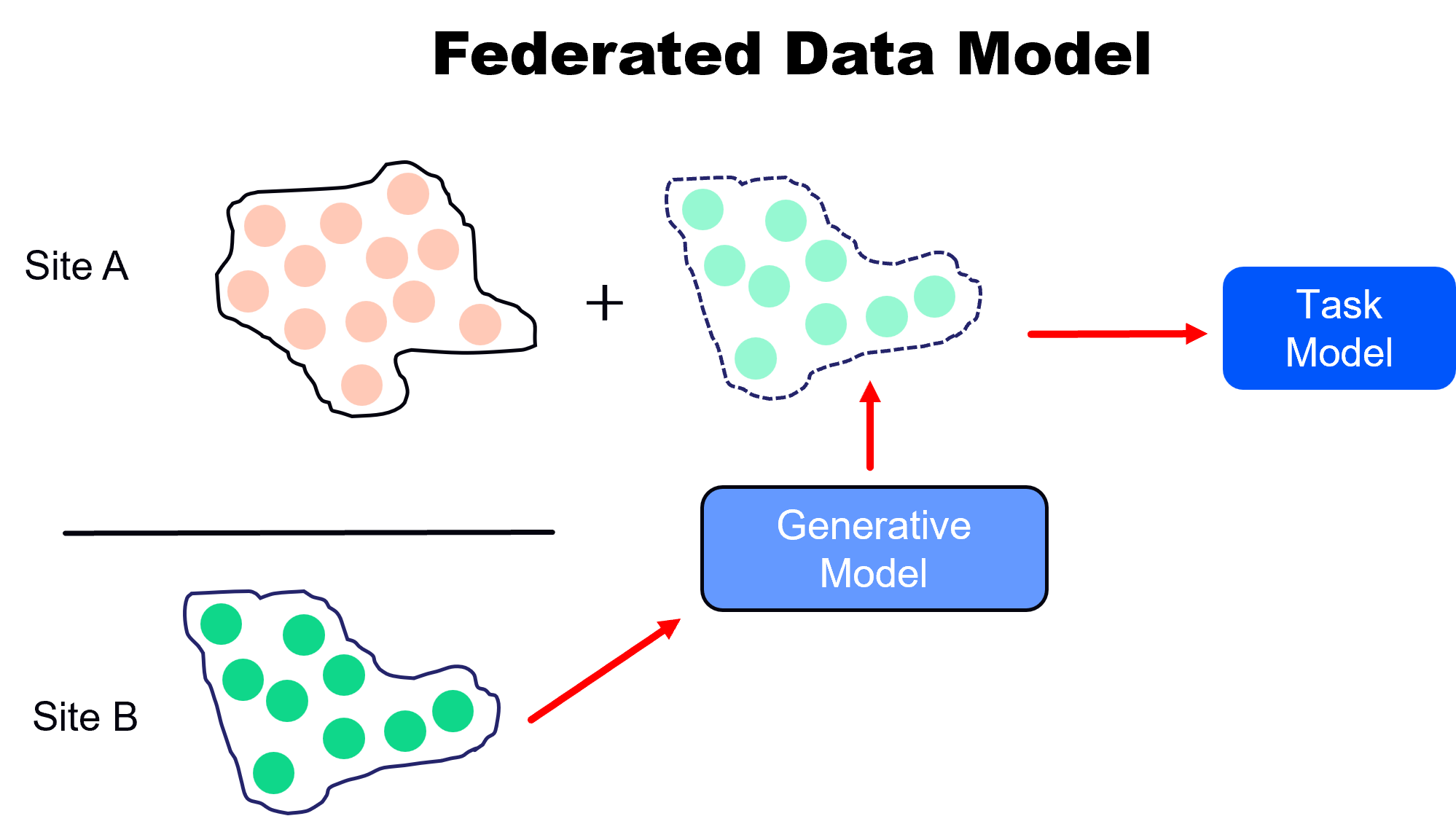}
\caption{\label{fig:method}Illustration of the proposed Federated Data Model (FDM).}
\end{figure}
\subsection{Federated Data Model (FDM)}
% In FDM, a data model based on neural network is trained to learn the underlying data distribution from the available training data at one site. The data model can then generate synthetic data which can be considered as a sample from the local data distribution. Instead of sharing the actual data, the trained data model can then be transferred and shared across different sites. For a site that receives the trained data model, it can use the data model to generate synthetic data that represent remote data distribution. And a targeted task model (such as segmentation) can be trained using a combination of actual local data and synthesized remote data.
The Federated Data Model (FDM) involves training a neural network-based data model to capture the local data distribution at a specific site. This model can generate synthetic data, effectively mirroring samples from the site's unique data distribution. Rather than transferring actual data across sites, the trained model is shared, enabling the recipient site to produce synthetic data reflective of another site's data distribution. This synthetic data, in conjunction with the recipient site's real data, facilitates the training of task-specific models (e.g., for segmentation), thereby harnessing both local and synthesized remote data distributions. Fig~\ref{fig:method} shows an illustration of FDM.

\subsection{Experiments}
Our experimental framework employs medical image segmentation, specifically focusing on segmenting the left ventricle myocardium (MYO) in T1 mapping cardiac magnetic resonance (CMR) images. These images were acquired under varying CMR scanning protocols across two distinct hospitals, labeled as Hospital A and Hospital B. Hospital A contributed approximately 750 slices from around 290 patients, and Hospital B provided about 1900 slices from roughly 250 patients, with each institution offering manually annotated slices.

Data from each site were partitioned into training, testing, and validation sets at a 60/20/20 ratio, ensuring patient-level separation. A UNet-based segmentation model \cite{unet} and a DDPM-based diffusion model \cite{ddpm} were trained on the training dataset from each site. The diffusion process was tailored to generate T1 maps with the left ventricle myocardium positioned as per the given segmentation masks.

The segmentation model's performance was initially assessed on each site's validation set—referred to as the "local" experiment. Subsequently, the diffusion model was transferred to the alternate site to produce synthetic T1 maps, conditioned on local segmentation masks. A segmentation model was then trained on this enriched dataset, combining local real and remote synthetic data, in what we designate as the "remote augment" experiment. 

\begin{figure}
\centering
\includegraphics[width=0.75\linewidth]{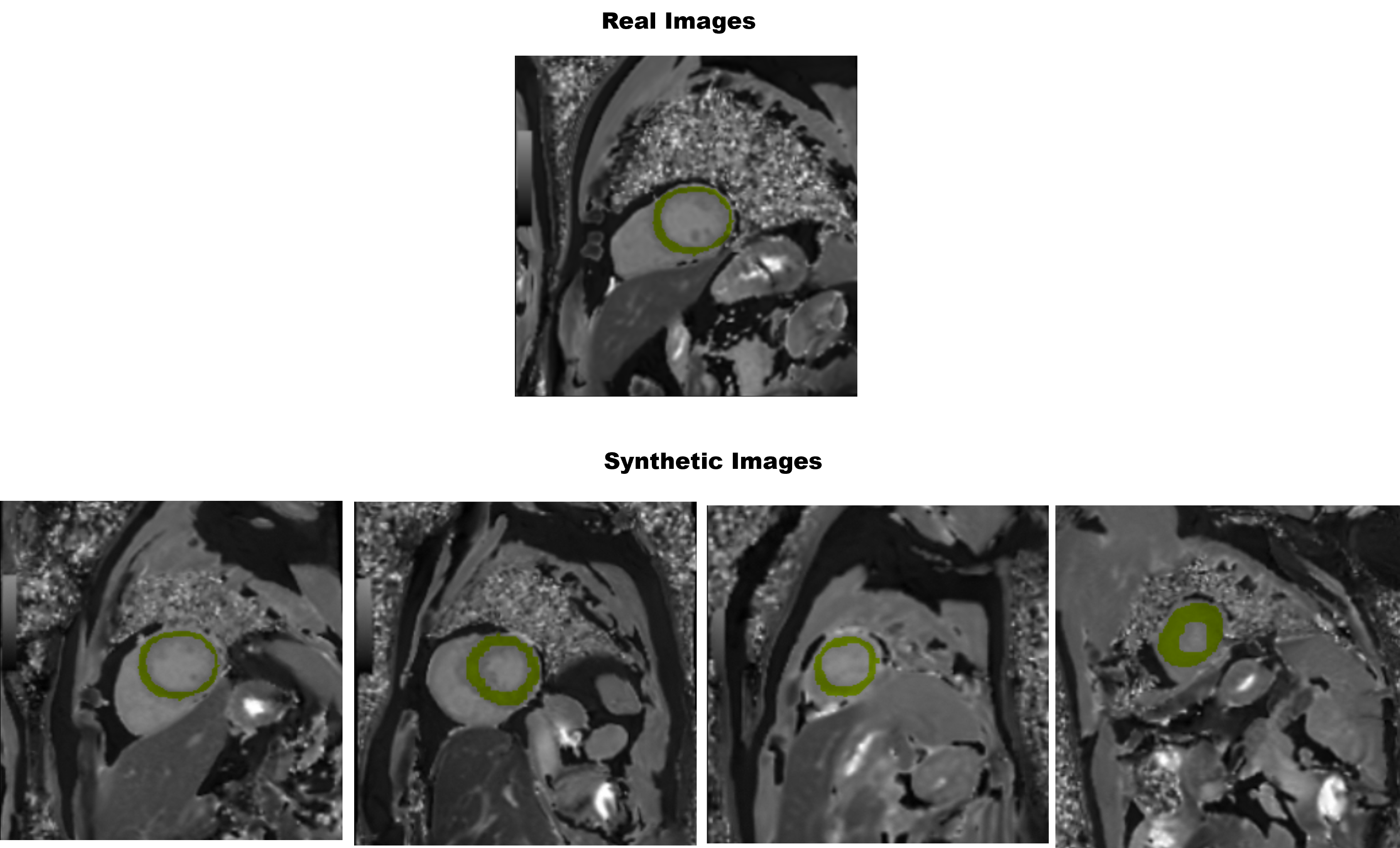}
\caption{\label{fig:example}Example real and synthetic images.}
\end{figure}

\section{Results}
\begin{figure}
\centering
\includegraphics[width=0.75\linewidth]{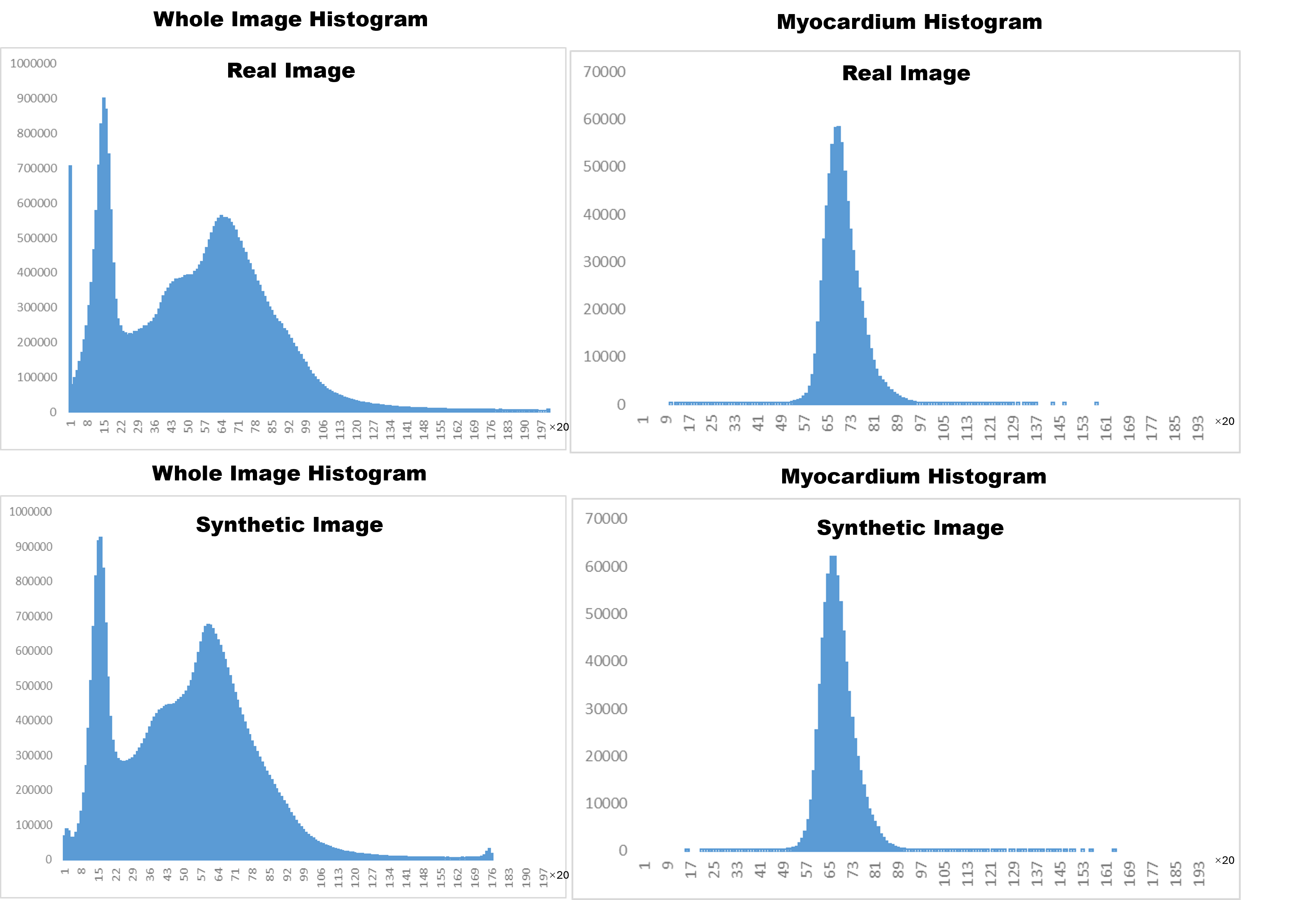}
\caption{\label{fig:hist}Histogram of images and myocardium regions from both real and synthetic data.}
\end{figure}

Example real and synthetic images are shown in Fig~\ref{fig:example}. 
Segmentation results measured by DICE scores are summarized in Table~\ref{tab:dice}. syn.B represents the synthetic data generated using the diffusion model trained on hospital B data.  
Histograms of the real and the synthetic images are shown in Fig~\ref{fig:hist} where both the whole image and the myocardium region, which is the segmentation target region were analyzed.

\section{Discussions}
Initially, when segmentation models were trained exclusively with data from their respective local datasets, the models achieved satisfactory performance on local test sets (Table~\ref{tab:dice}, row 1). However, there was a significant decline in accuracy when these models were applied to test sets from the remote site (Table~\ref{tab:dice}, row 2), indicating a pronounced challenge in model generalization across different data distributions.

Upon integrating the Federated Data Model (FDM) approach, where models were trained using synthetic data generated to represent the remote site's data distribution, we observed a notable improvement in model performance on remote test sets (Table~\ref{tab:dice}, row 3). This enhancement is captured in the comparison of results before and after applying FDM, highlighting the effectiveness of our method in addressing the domain shift problem.

Interestingly, the application of FDM not only preserved but slightly improved the model's accuracy on the original local test data (Table~\ref{tab:dice}). This improvement could be attributed to the augmented size of the training dataset provided by the inclusion of synthetic data, suggesting that our approach not only aids in overcoming domain shift challenges but also contributes positively to model robustness and generalizability.

% The results are reported in Table~\ref{tab:dice}. When the segmentation was trained on local data set only, although the local test performance can be quite satisfying (row 1), the remote test performance has a huge drop (row 2). When the model was trained using FDM, the performance has greatly improved (row 3). And the FDM did not destroy the model's accuracy on the original local data (row 4), even with slightly improved performance which may be due to augmented training data size. 

\section{Conclusions}
In this study, we have proposed a method Federated Data Model (FDM) to train accurate and privacy-respecting AI models across different locations. The method was validated on real medical application and the results demonstrate the feasibility and efficacy of applying our method across multiple sites, offering a promising avenue for developing robust models in privacy-sensitive and regulated environments.

\begin{table}
\centering
\begin{tabular}{c|c|c}
Train Data Source & Test Data Source & DICE \\\hline
Hospital A & Hospital A & 0.889 \\
Hospital A & Hospital B & 0.696 \\
Hospital A + syn.B & Hospital B & 0.820 \\
Hospital A + syn.B & Hospital A & 0.892
\end{tabular}
\caption{\label{tab:dice} Dice scores}
\end{table}

\bibliographystyle{splncs04}
\bibliography{sample}

\end{document}